\documentclass[journal]{IEEEtran}

\usepackage{latexsym,,amssymb,amsmath,graphicx,epsf,cite,bbm,float}
\usepackage{ifpdf}
\usepackage{epstopdf}

\usepackage{algorithm,algorithmic,hyperref}
\usepackage{amsmath,amssymb,bm}
\usepackage{amsfonts,dsfont,color,bbm,subcaption}

\newcommand\numberthis{\addtocounter{equation}{1}\tag{\theequation}}

\newcommand{\be}{\begin{equation}}
\newcommand{\ee}{\end{equation}}
\newcommand{\bea}{\begin{eqnarray}}
\newcommand{\eea}{\end{eqnarray}}

\newcommand{\MB}{\left[\begin{array}}
\newcommand{\ME}{\end{array}\right]}

\newcommand{\ei}{\end{itemize}}
\newcommand{\bi}{\begin{itemize}}

\usepackage{amsmath,amsthm}

\newcommand{\E}{\mathbb{E}}
\newcommand\Tau{\mathcal{T}}

\newtheorem{theorem}{Theorem}

\newtheorem{lemma}[]{Lemma}

\newtheorem{proposition}[]{Proposition}

\newtheorem{corollary}[]{Corollary}

\newtheorem{remark}[]{Remark}

\newtheorem{example}[]{Example}

\begin{document}

\title{Second Order Regret Bounds Against Generalized Expert Sequences under Partial Bandit Feedback} 
\author{Kaan Gokcesu, Hakan Gokcesu}
\maketitle

\flushbottom

\begin{abstract}
	We study the problem of expert advice under partial bandit feedback setting and create a sequential minimax optimal algorithm. Our algorithm works with a more general partial monitoring setting, where, in contrast to the classical bandit feedback, the losses can be revealed in an adversarial manner. Our algorithm adopts a universal prediction perspective, whose performance is analyzed with regret against a general expert selection sequence. The regret we study is against a general competition class that covers many settings (such as the switching or contextual experts settings) and the expert selection sequences in the competition class are determined by the application at hand. Our regret bounds are second order bounds in terms of the sum of squared losses and the normalized regret of our algorithm is invariant under arbitrary affine transforms of the loss sequence. Our algorithm is truly online and does not use any preliminary information about the loss sequences. 
\end{abstract}

\section{Introduction}\label{sec:intro}

In machine learning literature \cite{mohri2018foundations,jordan2015machine}, the field of online learning \cite{shalev2011online} is heavily studied in a myriad of fields from control theory \cite{tnnls3} and computational learning theory \cite{comp2,comp1} to decision theory \cite{freund1997,tnnls4}. Especially, algorithms pertaining to the universal prediction perspective \cite{merhav} have been utilized in many signal processing applications \cite{gokcesu2021nonparametric,signal1,moon,sw5}, and in sequential estimation/detection problems \cite{gHierarchical,singer,gokcesu2021optimally,singer2} such as density estimation or anomaly detection \cite{willems,gAnomaly,coding2,coding1,gDensity}. Some of its most popular applications are in multi-agent systems \cite{tekin2014distributed,vanli,sw1,gokcesu2020recursive,gokcesu2021optimal} and more prominently in reinforcement learning problems \cite{bandit1,auerExp,vural2019minimax,audibert,bandit2,neyshabouri2018asymptotically,exptrade,tekin2,gBandit,auerSelf,reinInt,auer}, where the famous exploration and exploitation trade off is commonly encountered \cite{cesa-bianchi}. 

In many sequential decision making problems, the setting of imperfect feedback is most generally encompassed by the framework of partial monitoring \cite{bartok2014partial}; where the goal is to sequentially select one of existing $M$ actions (e.g., expert advice and parallel running algorithms) and minimize (or maximize) some loss (or reward) whilst observing the outcomes of the actions in a limited partial manner \cite{cesabook}. Various interesting problems can be modeled by the partial monitoring framework, such as dynamic pricing \cite{kleinberg2003value}, label efficient prediction \cite{cesa2005minimizing}, linear/convex optimization with full/bandit feedback \cite{zinkevich2003online,flaxman2004online,abernethy2008competing}, the dark pool problem \cite{agarwal2010optimal} and the popular multi-armed bandit problem \cite{bandit2,cesa-bianchi,audibert,auer,auerExp,banditTNN,zheng,gBandit}, which is (in some sense) a limited feedback version of the traditional prediction with expert advice \cite{littlestone1994,signal2,signal1,singer,moon,singer2}. This setting is applicable in a wide range of problems from recommender systems \cite{tang2014ensemble,li2011unbiased,luo2015nonnegative,tekin2014distributed}, cognitive radio \cite{gai2010learning,lai2008medium} and clinical trials \cite{hardwick1991bandit} to online advertisement \cite{li2010contextual}. We study this problem in an online setting, where we sequentially operate on an adversarial observation stream \cite{huang2011adversarial} and investigate the problem from a competitive algorithm perspective \cite{merhav,vural2019minimax,vovk,littlestone1994,neyshabouri2018asymptotically,vovk1998,gokcesu2020recursive}. 

The competitive perspective is achieved by the utilization of regret bounds. In this perspective, the performance of our selections are compared against a competition class of expert selection strategies. Given a loss sequence, the aim of an algorithm is to achieve a total loss that is as good as the total loss of the competitor selections (e.g., for fixed competitions, we compare against the single expert selection with the best cumulative loss) \cite{cesa2007}. The difference between the total loss of our selections and the best strategy is called regret \cite{cesabook}. 

The regret bounds of the partial monitoring problems are ongoing research subjects. A special case of partial monitoring, the adversarial multi-armed bandit problem, has a regret lower bound of $\Omega(\sqrt{T})$ in a $T$ round game against the best fixed selection \cite{cesa-bianchi}. This regret lower bound implies a minimax bound of $\Omega(\sqrt{W_*T})$, where $W_*$ is the sequence complexity of an expert selection strategy in an arbitrary competition class \cite{gBandit,gokcesu2020generalized,gokcesu2021generalized}. This complexity can be dependent on either the number of switches in the sequence \cite{cesa-bianchi,auer,auerExp,audibert,doubling_trick}, the number of contextual regions \cite{willems1995context,sadakane2000implementing,willems1996context,csiszar2006context,dumont2014context,kozat2007universal,vanli2014comprehensive}, or any other complexity definition that implies a prior on the expert sequences \cite{comp2,gokcesu2020generalized,gokcesu2021generalized}. With some alterations, there exist state of the art algorithms \cite{gokcesu2020generalized,gokcesu2021generalized} that can achieve a regret bound of $O(\sqrt{WT})$ when competing against the expert sequence with complexities upper bounded by $W$.
In a more general partial monitoring setting we have a regret lower bound of $\Omega(T^{\frac{2}{3}})$ in a $T$ round game against the best fixed selection \cite{cesa2006regret}. Hence, a regret bound of $O(W^{\frac{1}{3}}T^{\frac{2}{3}})$ is implied, which is addressed in this work.

Moreover, it is a popular topic of study to create fundamental, translation/scale-free regret bounds. Against fixed competitions, the algorithm of exponentially weighted averaging \cite{littlestone1994,vovk1998} provides a zeroth order regret bound. In one-sided games (i.e., all losses have the same sign), \cite{freund1997} showed that the algorithm of \cite{littlestone1994} obtains a first order regret bound. A direct study on the signed games in the work of \cite{allenberg2004} finds that weighted majority achieves the first order regret without a need for one-sided losses. Although these approaches are scale equivariant, they are not translation invariant.
The work in \cite{cesa2007} solves these shortcomings by creating second order regret bounds that requires no a priori knowledge. Their regret bounds are translation, scale invariant and also parameter-free. However, their competition class is limited and focused on best fixed selection. 
The approaches in \cite{gokcesu2020generalized,gokcesu2021generalized} address this issue by extending the second order regret bounds to generalized competition classes that are able to arbitrarily compete against different choices of competitions in the problem of prediction with expert advice and multi-armed bandits. In this work, we aim to extend these regret results to the harder problem of prediction in the presence of partial limited feedback.

\section{Problem Description}\label{sec:problem}
In this work, we study the mixture of experts problem under partial monitoring. We have $M$ experts such that $m\in\{1,\ldots,M\}$ and randomly select one of them at each round $t$ according to our selection probabilities \cite{gokcesu2020generalized,gokcesu2021generalized}
\begin{align}
	q_t\triangleq[q_{t,1},\ldots,q_{t,M}].\label{eq:qt}
\end{align}
Based on our online selection $\{i_t\}_{t\geq1},\enspace i_t\in\{1,2,\ldots,M\}$, we incur the loss of the selected expert $\{{l_{t,i_t}}\}_{t\geq1}$,
where we do not assume anything about the losses before selecting our expert at time $t$. Because of the partial monitoring setting, we observe the losses $\{{l_{t,{m}}}\}_{t\geq1}$ randomly (i.e., they may be hidden).
Let $\mathcal{P}_t(m,m')$ be the possibility of observing the loss of $m^{th}$ expert when $(m')^{th}$ expert is selected at time $t$.

\begin{remark}
	When $\mathcal{P}_t(m,m')=1, \forall m,m'$; we have the full feedback setting.
\end{remark}

For a more general analysis, we assume that $\mathcal{P}_t(\cdot,\cdot)$ can be arbitrary and can even change in each round $t$. Therefore, it can even be chosen in an adversarial manner. However, for observability of every expert, we make the following notion:
\begin{align}
	\sum_{m'}\mathcal{P}_t(m,m')=1, \forall m
\end{align} 

\begin{example}
	In the special case of when $\mathcal{P}_t(m,m')=1$ if $m=m'$ and $0$ otherwise; we have the classic bandit feedback.
\end{example}

In a $T$ round game, we define $I_T$ as the row vector containing the user selections up to time $T$ and its loss sequence $L_{I_T}$, i.e.,
\begin{align}
	I_T=[i_1,\ldots,i_T],&&L_{I_T}=[l_{1,{i_1}},\ldots,l_{T,i_T}].
\end{align}
Similarly, we define $S_T$ as the row vector representing a deterministic expert selection sequence of length $T$ and its loss sequence $L_{S_T}$, i.e.,
\begin{align}
	S_T=[s_1,\ldots,s_T],&&	L_{S_T}=[l_{1,{s_1}},\ldots,l_{T,s_T}].\label{eq:St}
\end{align}
such that each $s_t\in\{1,2,\ldots,M\}$ for all $t$. In the rest of the paper, we refer to each such deterministic expert selection sequence, $S_T$, as a competition. 
We denote the cumulative losses at time $T$ of $I_T$ and $S_T$ by 
\begin{align}
	{C_{I_T}=\sum_{t=1}^T l_{t,i_t}},&&{C_{S_T}=\sum_{t=1}^T l_{t,s_t}}.
\end{align}
Since we assume no statistical assumptions on the loss sequence, we define our performance with respect to a competition $S_T$ that we want to compete against.
We use the notion of universal regret to define our performance against any competition $S_T$ as
\begin{align}
R_{S_T}&\triangleq C_{I_T}-C_{S_T}=\sum_{t=1}^T l_{t,i_t}-\sum_{t=1}^T l_{t,s_t},\label{RT1}
\end{align}
where we denote the regret accumulated in $T$ rounds against $S_T$ as $R_{S_T}$. Our goal is to create an algorithm with minimax optimal expected regret bounds that are translation and scale free against $S_T$ that depends on how hard it is to learn the competition $S_T$ \cite{gokcesu2020generalized,gokcesu2021generalized}.

\section{The Framework}\label{sec:method}
	Our framework starts by assigning a weight $w_{S_t}$ to each competition $S_t$, which are implicitly calculated with equivalence classes (same class sequences are updated together) \cite{gBandit,gokcesu2020generalized,gokcesu2021generalized}.
	We define the equivalence classes with
	\begin{align}
		\lambda_t=[m, \ldots],\label{lamt}
	\end{align}
	where the first parameter $\lambda_t(1)$ is expert selection $m$ at time $t$. Together with the omitted parameters in \eqref{lamt}, $\lambda_t$ determines the competitions that are included in that equivalence class, which consists of the competitions $S_t$ whose behavior match with the parameters $\lambda_t$. The parameters of $\lambda_t$ determine the number of equivalence classes and how many competitions each class represents. We define $\Omega_t$ as the set of all $\lambda_t$ as $\lambda_t\in\Omega_t, \enspace\forall\lambda_t.$
	$\Omega_t$ may not represent all possible sequences at time $t$, but instead the sequences of interest (competition). We define $\Lambda_t$ as the parameter sequence up to $t$ as
	\begin{align}
		\Lambda_t\triangleq\{\lambda_1,\ldots,\lambda_t\},\label{Lamt}
	\end{align} 
	where each sequence $S_t$ will correspond to only one $\Lambda_t$. 
	We define $w_{\lambda_t}$ as the weight of the equivalence class parameters $\lambda_t$ at time $t$. The weight of an equivalence class is simply the summation of the implicit weights of the sequences whose behavior conforms with its class parameters $\lambda_t$, such that
	\begin{align}
		w_{\lambda_t}=\sum_{F_\lambda(S_t)=\lambda_t}^{}w_{S_t},\label{wlt}
	\end{align}
	where $F_\lambda(\cdot)$ is the mapping from sequences $S_t$ to the auxiliary parameters $\lambda_t$.
	Similar to \cite{gokcesu2020generalized,gokcesu2021generalized}, we update the weights $w_{\lambda_t}$ using the following two-step approach. First, we define an intermediate variable $z_{\lambda_t}$ (which incorporates the exponential update as in the exponential weighting \cite{cesabook}) such that
	\begin{align}
		z_{\lambda_t}\triangleq w_{\lambda_t}e^{-\eta_{t-1}\phi_{t,\lambda_t(1)}},\label{zlt}
	\end{align}
	where $\phi_{t,\lambda_t(1)}$ is a measure of the performance $(\lambda_t(1))^{th}$ expert at time $t$, which we discuss more in the next section.
	Secondly, we create a probability sharing network among the equivalence classes (which also represents and assigns a weight to every individual sequence $S_t$ implicitly) at time $t$ as
	\begin{align}
		w_{\lambda_{t+1}}=\sum_{\lambda_t\in\Omega_t}\Tau(\lambda_{t+1}|\lambda_t)z_{\lambda_t}^{\frac{\eta_t}{\eta_{t-1}}},\label{wlt+}
	\end{align}
	where $\Tau(\lambda_{t+1}|\lambda_t)$ is the transition weight from the class parameters $\lambda_t$ to $\lambda_{t+1}$ such that $\sum_{\lambda_{t+1}\in\Omega_{t+1}}\Tau(\lambda_{t+1}|\lambda_t)=1$ (which is a probability distribution itself). The power normalization on $z_{\lambda_{t}}$ is necessary for adaptive learning rates \cite{gokcesu2020generalized,gokcesu2021generalized}.
	Using $w_{\lambda_t}$, we construct the expert weights as
	\begin{align}
		w_{t,m}=\sum_{\lambda_t(1)=m}^{} w_{\lambda_t}.\label{wmt2}
	\end{align}
	The probabilities are constructed by normalization, i.e.,
	\begin{align}
	p_{t,m}=\frac{w_{t,m}}{\sum_{m'}w_{t,m'}},\label{pmt}
	\end{align}
	and selection probabilities $q_{t,m}$ are given by mixing $p_{t,m}$ with a uniform distribution (similarly to \cite{gokcesu2020generalized,gokcesu2021generalized}) as 
	\begin{align}
		q_{t,m}=(1-\epsilon_t)p_{t,m}+\epsilon_t\frac{1}{M},\label{qtm}
	\end{align}
	where $\epsilon_t$ is a time dependent uniform mixing parameter.

\section{Algorithm Design and Regret Bounds}\label{sec:regret}
	In this section, we study the performance of our algorithm. We first provide a summary of some important notations and definitions, which will be heavily used. Then, we study the regret bounds by successively designing the learning rates $\eta_t$, the performance measures $\phi_{t,m}$ and the uniform mixture coefficients $\epsilon_{t}$.
	
	\subsection{Preliminaries}
	
	Before starting our design and analysis, we provide some relevant definitions and notations below.
	\begin{enumerate}
		\item $q_{t,m}$ is the probability of selecting $m$ at $t$ as in \eqref{qtm}.
		\item $\E_{f_{t,m}}[x_{t,m}]$ is the convex sum of $x_{t,m}$ with the coefficients $f_{t,m}$, i.e., $\sum_{m=1}^{M}f_{t,m}x_{t,m}$.
		\item $\E[x]$ is the expectation of $x$ when user selection $\{i_t\}_{t\geq 1}$ is drawn from $\{q_{t,m}\}_{t\geq 1}$.
		\item $\mathbbm{1}_{t,m}$ is the indicator function for observation, i.e., $\mathbbm{1}_{t,m}$ is $1$ if the loss of $m$ is observed at time $t$ and $0$ otherwise.
		\item $\eta_t$ is the learning rate used in \eqref{zlt}.
		\item $\phi_{t,m}$ is the performance metric used in \eqref{zlt}.
		\item $d_t\triangleq \enspace\max_m\phi_{t,m}-\min_m\phi_{t,m}$.
		\item $v_t\triangleq \enspace\E_{p_{t,m}}\phi_{t,m}^2$.
		\item $D_t\triangleq\max_{1\leq t' \leq t}d_t,$.
		\item $V_t\triangleq\sum_{t'=1}^t v_t$.
		\item $\log(\cdot)$ is the natural logarithm. 
		\item $\lambda_t$ is an equivalence class parameter at time $t$ as in \eqref{lamt}.
		\item $\Omega_t$ is the set of all $\lambda_t$ at time $t$.
		\item $\Lambda_T\triangleq\{\lambda_t\}_{t=1}^T$ as in \eqref{Lamt}.
		\item $z_{\lambda_{t}}$ is as in \eqref{zlt}.
		\item $\Tau(\cdot|\cdot)$ is the transition weight used in \eqref{wlt+}.
		\item $\Tau(\{\lambda_t\}_{t=1}^T)\triangleq\prod_{t=1}^T\Tau(\lambda_t|\lambda_{t-1})$.		
		\item $W(\Lambda_T)\triangleq \log(\max_{1\leq t\leq T}|\Omega_{t-1}|)-\log(\Tau(\Lambda_T))$, which corresponds to the complexity of a competition.
\end{enumerate}
	
	We start our algorithmic design similar to \cite{gokcesu2020generalized,gokcesu2021generalized}, where there exist three design components in our framework. Specifically, these are the learning rates $\eta_t$, performance measures $\phi_{t,m}$ and the uniform mixture $\epsilon_t$. We will gradually construct these in our analysis.
	
	\begin{itemize}
		\item We start by setting the following learning rates \cite{gokcesu2021generalized}
		\begin{align}
			\eta_t=\frac{\gamma}{\sqrt{V_t+D_t^2}},\label{etat}
		\end{align}
		which are non-increasing and $\gamma$ is a user-set parameter. 
		
		\item We set the performance measure $\phi_{t,m}$ as
		\begin{align}
			\phi_{t,m}=&\begin{cases}
				\frac{l_{t,m}-\psi_t}{o_{t,m}}, &\mathbbm{1}_{t,m}=1\\
				0, &\text{ otherwise}
			\end{cases},\label{phitm}
		\end{align}
		\begin{itemize}
			\item $l_{t,m}$ is the loss of the $m^{th}$ expert, 
			\item $o_{t,m}$ is the observation probability of the expert $m$ at time $t$, i.e.,
			\begin{align}
				o_{t,m}=\sum_{m'}\mathcal{P}_t(m,m')q_{t,m'}\label{otm}
			\end{align} 
			\item $\psi_t$ is the minimum loss observed so far, i.e.,
			\begin{align}
				\psi_t=&\min (\psi_{t-1},\min_{m: \mathbbm{1}_{t,m}=1}l_{t,m}).\label{psit}
			\end{align}
		\end{itemize}
	\end{itemize}
	
	\subsection{Analysis of Performance Measure}
	The performance analysis starts similarly to \cite{gokcesu2020generalized,gokcesu2021generalized}, where we have
	
	\begin{lemma}\label{thm:bound}
		We have
		\begin{align*}
			\sum_{t=1}^T\left(\E_{p_{t,m}}\phi_{t,m}-\phi_{t,\lambda_t(1)}\right)\leq& \frac{1}{2}\sum_{t=1}^T\eta_t\E_{p_{t,m}}\phi_{t,m}^2\\
			&+\sum_{t=1}^T\left(1-\frac{\eta_t}{\eta_{t-1}}\right)d_t\\
			&+\frac{\log(\max_{1\leq t\leq T}|\Omega_{t-1}|)}{\eta_{T-1}}\\
			&-\frac{1}{\eta_{T-1}}\log(\Tau(\Lambda_T)),
		\end{align*}
		where $\Tau(\Lambda_T)=\Tau(\{\lambda_t\}_{t=1}^T)$; $\phi_{t,m}\geq 0$, for all $t,m$; $\eta_t$ is non-increasing with $t$.
		\begin{proof}
			The proof follows from \cite{gokcesu2020generalized,gokcesu2021generalized}, where we use the inequality $e^{-x}\leq 1-x+\frac{1}{2}x^2$ for $x\geq 0$.
		\end{proof}
	\end{lemma}
	\autoref{thm:bound} provides us an upper bound on the cumulative difference on the performance variable $\phi_{t,m}$ in terms of the learning rates $\eta_t$ and the performance measures $\phi_{t,m}$. Next, we have some result from the selection of the learning rates $\eta_t$ in \eqref{etat}. 
	
	\begin{lemma}\label{thm:ntphi2}
		When using the learning rates in \eqref{etat}, we have
		\begin{align*}
			\frac{1}{2}\sum_{t=1}^T\eta_t\E_{p_{t,m}}\phi_{t,m}^2\leq \gamma\sqrt{V_T},
		\end{align*}
	where $\gamma$ is a user-set parameter.
		\begin{proof}
			Proof is from \cite{gokcesu2021generalized}.
		\end{proof}
	\end{lemma}

	\begin{lemma}\label{thm:ntdt}
		When using the learning rates in \eqref{etat}, we have
		\begin{align*}
			\sum_{t=1}^{T}\left(1-\frac{\eta_t}{\eta_{t-1}}\right)d_t\leq&\sqrt{V_T+D_T^2}.
		\end{align*}
		\begin{proof}
			Proof follows \cite{gokcesu2021generalized}.
		\end{proof}
	\end{lemma}

	\begin{lemma}\label{thm:bound2}
		When using the learning rates in \eqref{etat}, we have
		\begin{align*}
		\sum_{t=1}^T\E_{p_{t,m}}\phi_{t,m}-\phi_{t,s_t}
		\leq&\frac{W(\Lambda_T)+\gamma}{\gamma}\sqrt{V_T+D_T^2}+{\gamma\sqrt{V_T}},
		\end{align*}
		where $\gamma$ is a user-set parameter and $s_t\triangleq \lambda_{t}(1)$.
		\begin{proof}
			Similarly with \cite{gokcesu2021generalized}, the proof comes from combining \autoref{thm:bound} with \autoref{thm:ntphi2} and \autoref{thm:ntdt} and the definition of $W(\Lambda_T)$.
		\end{proof}
	\end{lemma}
	\autoref{thm:bound2} provides us an upper bound on the cumulative difference on the performance variable $\phi_{t,m}$ in terms of the isolated parameter $\gamma$ which needs to be set at the beginning. However, this does not invalidate the fact that our algorithm is truly online since $\gamma$ will be straightforwardly set based on the competition class, which is available at the start of the design of our algorithm. 
	
	\subsection{Second Order Regret}
	Here, we construct the second order bounds for the regret of our algorithm. To do this, we first need to create the regret with respect to the expert selection probabilities $q_{t,m}$ instead of the algorithmic probabilities $p_{t,m}$.
	\begin{proposition}\label{thm:EpEq}
		When $q_{t,m}$ is constructed by mixing $p_{t,m}$ with uniform probabilities, we have
		\begin{align*}
			\E_{p_{t,m}}\phi_{t,m}=\E_{q_{t,m}}\phi_{t,m}+\epsilon_{t}\left[\E_{p_{t,m}}\phi_{t,m}-\E_u\phi_{t,m}\right],
		\end{align*}
		where $\E_u$ is expectation with uniform probability.
		\begin{proof}
			The proof is straightforward from \eqref{qtm}.
		\end{proof}
	\end{proposition}`	
	Next, we provide a result for the expectation of the performance measures and show its relation to the losses.
	\begin{proposition}\label{thm:EphiEpsi}
		For the expectation of the performance variable $\phi_{t,m}$ in \eqref{phitm}, we have
		\begin{align*}
			\E[\phi_{t,m}]=\E[l_{t,m}]-\E[\psi_t|\mathbbm{1}_{t,m}=1].
		\end{align*}
		\begin{proof}
			We have
			\begin{align}
				\E_{}[\phi_{t,m}]=&\E[\E[\phi_{t,m}|\mathbbm{1}_{t,m}]],\\
				=&o_{t,m}\E\left[\frac{l_{t,m}-\psi_t}{o_{t,m}}|\mathbbm{1}_{t,m}=1\right]\\
				=&\E[l_{t,m}]-\E[\psi_t|\mathbbm{1}_{t,m}=1],
			\end{align}
		which concludes the proof.
		\end{proof}
	\end{proposition}

	Next, we relate the regret with respect to the algorithmic probabilities to the regret with respect to the expert selection probabilities.
	\begin{lemma}\label{thm:ERpERq}
		We have
		\begin{align*}
			&\E\left[\sum_{t=1}^T\E_{p_{t,m}}[\phi_{t,m}]-\phi_{t,s_t}\right]\\
			&\geq\E\left[\sum_{t=1}^T(\E_{q_{t,m}}[l_{t,m}]-l_{t,s_t})\right]-\sum_{t=1}^{T}\E[\E_{q_{t,m}}\E[\psi_t|\mathbbm{1}_{t,m}=1]]\\
			&\enspace\enspace+\sum_{t=1}^T\E[\psi_t|\mathbbm{1}_{t,s_t}=1]-\sum_{t=1}^{T}\epsilon_t(A-B),
		\end{align*}
		where $[B,A]$ is the range of losses $l_{t,m}$.
		\begin{proof}
			The proof comes from combining \autoref{thm:EpEq} and \autoref{thm:EphiEpsi}.
		\end{proof}
	\end{lemma}
	
	By utilizing these results, we have the following second order regret bound.
	\begin{theorem}\label{thm:ERST}
		The expected regret of our algorithm is given by
		\begin{align*}
			\E[R_{S_T}]\leq& \frac{W(\Lambda_T)+\gamma}{\gamma}\sqrt{\E[V_T]+\E[D_T^2]}+{\gamma\sqrt{\E[V_T]}}\\
			&+\sum_{t=1}^{T}\epsilon_t(A-B)+\E[\Psi_T],
		\end{align*}
		where $$\E[\Psi_T]\triangleq \sum_{t=1}^{T}\E[\E_{q_{t,m}}\E[\psi_t|\mathbbm{1}_{t,m}=1]]-\sum_{t=1}^T\E[\psi_t|\mathbbm{1}_{t,s_t}=1].$$
		\begin{proof}
			The proof comes from \autoref{thm:bound2} and \autoref{thm:ERpERq}.
		\end{proof}
	\end{theorem}
	
	\subsection{Bounding \texorpdfstring{$V_T$}{VT} and \texorpdfstring{$D_T$}{DT}}
	To bound the expected regret, we need to bound both of the expectations of $V_t$ and $D_t^2$. For this, we first utilize the following result.
	\begin{proposition}\label{thm:otmbound}
		The observation probabilities $o_{t,m}$ are lower bounded as
		\begin{align*}
			o_{t,m}\geq \frac{\epsilon_t}{M}, &&\forall t,m.
		\end{align*}
		\begin{proof}
			We have $q_{t,m}\geq \frac{\epsilon_{t}}{M}$ from \eqref{qtm}. Since $\sum_{m'}\mathcal{P}(m,m')=1$, we have $o_{t,m}\geq \frac{\epsilon_{t}}{M}$, which concludes the proof.
		\end{proof}
	\end{proposition}
	
	Then, we create a bound on the expectation of $V_T$.
	\begin{lemma}\label{thm:EVT}
		For $\phi_{t,m}$ as in \eqref{phitm}, $\psi_t$ as in \eqref{psit} and $o_{t,m}$ as in \eqref{otm}, the expectation of $V_T$ is bounded as follows:
		\begin{align*}
			\E[V_T]&\leq M(A-B)^2\sum_t \frac{1}{\epsilon_t},
		\end{align*}
		where $[B,A]$ is the range of losses $l_{t,m}$ for all $t,m$.
		\begin{proof}
			From the definition of $v_t$, we have
			\begin{align}
				\E[v_t]&=\E\left[\sum_mp_{t,m}\frac{(l_{t,m}-\psi_t)^2}{o_{t,m}}\right],\\
				&\leq\E\left[\sum_m\frac{p_{t,m}}{o_{t,m}}(A-B)^2\right]\label{Epvt}
			\end{align}
			From \autoref{thm:otmbound}, we have
			\begin{align}
				\frac{1}{o_{t,m}}\leq \frac{M}{\epsilon_t}\label{1/o}
			\end{align}
			Thus, combining \eqref{Epvt} and \eqref{1/o} gives
			\begin{align}
				\E [v_t]&\leq\frac{M(A-B)^2}{\epsilon_t},
			\end{align}
			Thus,
			\begin{align}
				\E[V_T]&\leq M(A-B)^2\sum_t \frac{1}{\epsilon_t},
			\end{align}
			which concludes the proof.
		\end{proof}
	\end{lemma}
	
	Similarly, we also create a bound on the expectation of $D_T^2$.
	\begin{lemma}\label{thm:EDT}
		For $\phi_{t,m}$ as in \eqref{phitm}, $\psi_t$ as in \eqref{psit} and $o_{t,m}$ as in \eqref{otm}, the expectation of $D_T^2$ is bounded as follows:
		\begin{align}
			\E[D_T^2]\leq\frac{M^2(A-B)^2}{\epsilon_T^2}
		\end{align}
		where $[B,A]$ is the range of losses $l_{t,m}$ for all $t,m$.
		\begin{proof}
			The proof is straightforward from the definition of $D_T$
			\begin{align}
				D_T&=\max_{1\leq t \leq T}\frac{l_{t,i_t}-\psi_t}{o_{t,i_t}},\\
				&\leq\frac{M(A-B)}{\epsilon_T},
			\end{align}
			which concludes the proof.
		\end{proof}
	\end{lemma}

	\subsection{Bounding \texorpdfstring{$\psi_t$}{phit}}
	In this section, we construct the expected regret bound. To this end, we need to bound the expectation of the expression $\Psi_T$. To do this, we first bound the expectation of $\psi_t$.
	\begin{proposition}\label{thm:psit<t-1}
		We have
		\begin{align*}
			\E[\psi_t|\mathbbm{1}_{t,m}=b]\leq&\E[\psi_{t-1}],
		\end{align*}
		for any $b\in\{0,1\}$.
		\begin{proof}
			Since $\psi_t\leq \psi_{t-1}$ by definition, we have
			\begin{align}
				\E[\psi_t|\mathbbm{1}_{t,m}=b]\leq& \E[\psi_{t-1}|\mathbbm{1}_{t,m}=b],\\
				\leq&\E[\psi_{t-1}],
			\end{align}
			since $\psi_{t-1}$ is not a function of $\mathbbm{1}_{t,m}$. 
		\end{proof}
	\end{proposition}
	
	Next, we bound the cumulative sum of the expectation of $\psi_t$.
	\begin{lemma}\label{thm:Epsit}
	For $\phi_{t,m}$ as in \eqref{phitm}, $\psi_t$ as in \eqref{psit} and $q_{t,m}$ as in \eqref{qtm}, we have the following expectation result
	\begin{align*}
		\sum_{t=1}^{T-1}\E[\psi_t]\leq&\sum_{t=2}^{T-1}(Q_t-Q_{t-1}+1)\E[\psi_t|\mathbbm{1}_{t,s_t}=1]\\
		&+Q_1\E[\psi_1],
	\end{align*}
	for any $\{s_t\}_{t=1}^T$ expert selection sequence, where $[B,A]$ is the range of losses $l_{t,m}$ for all $t,m$ and $Q_t\triangleq\sum_{\tau=1}^{T-t}(1-\delta)^{\tau-1}$, where $\delta\leq o_{t,m}$ for all $t,m$.
	\begin{proof}
	For any $m$, we have
	\begin{align}
		\E[\psi_t]=&o_{t,m}\E[\psi_t|\mathbbm{1}_{t,m}=1]+(1-o_{t,m})\E[\psi_t|\mathbbm{1}_{t,m}=0],\\
		\leq&o_{t,m}\E[\psi_t|\mathbbm{1}_{t,m}=1]+(1-o_{t,m})\E[\psi_{t-1}],\\
		\leq&\delta\E[\psi_t|\mathbbm{1}_{t,m}=1]+\left(1-\delta\right)\E[\psi_{t-1}].
	\end{align}
	By the telescoping rule, we have
	\begin{align*}
		\E[\psi_t]\leq& \enspace\delta\E[\psi_t|\mathbbm{1}_{t,s_t}=1]\\
		&+(1-\delta)\delta\E[\psi_{t-1}|\mathbbm{1}_{t-1,s_{t-1}}=1]\\
		&+\ldots\\
		&+(1-\delta)^{t-2}\delta\E[\psi_2|\mathbbm{1}_{2,s_2}=1]\\
		&+(1-\delta)^{t-1}\E[\psi_1]\numberthis\label{Epsit},
	\end{align*}
	for any $\{s_\tau\}_{\tau=2}^t$ sequence, where $\delta\leq o_{t,m} ,\forall t,m$.
	By rearranging, we have
	\begin{align}
		\sum_{t=1}^{T-1}\E[\psi_t]\leq\sum_{t=2}^{T-1}\delta Q_t\E[\psi_t|\mathbbm{1}_{t,s_t}=1]+\sum_{t=1}^{T-1}(1-\delta)^{t-1}\E[\psi_1],
	\end{align}
	where $Q_t\triangleq\sum_{\tau=1}^{T-t}(1-\delta)^{\tau-1}$. Hence, $Q_{t-1}=(1-\delta)Q_t+1$ and $\delta Q_t=Q_t-Q_{t-1}+1$. Thus, we have
	\begin{align}
		\sum_{t=1}^{T-1}\E[\psi_t]\leq&\sum_{t=2}^{T-1}(Q_t-Q_{t-1}+1)\E[\psi_t|\mathbbm{1}_{t,s_t}=1]\nonumber\\
		&+Q_1\E[\psi_1],
	\end{align}
	which concludes the proof.
	\end{proof}
	\end{lemma}
	
	\subsection{Bounding \texorpdfstring{$\Psi_T$}{PhiT} and the Normalized Regret}
	We combine these results to bound the expression $\Psi_t$.
	\begin{lemma}\label{thm:EPsiT}
		We have
		\begin{align*}
			\E[\Psi_T]\leq\frac{1+\delta}{\delta}(A-B),
		\end{align*}
		where $[B,A]$ is the range of losses $l_{t,m}$ and $\delta\leq o_{t,m}$.
		\begin{proof}
			Using \autoref{thm:psit<t-1} and \autoref{thm:Epsit}, we have
			\begin{align}
				\sum_{t=1}^T(\E[\E&_{q_{t,m}}[\E[\psi_t|\mathbbm{1}_{t,m}=1]]-\E[\psi_t|\mathbbm{1}_{t,m}=1]])\nonumber\\
				\leq&\sum_{t=1}^{T-1}\E[\psi_t]-\sum_{t=1}^T\E[\psi_t|\mathbbm{1}_{t,s_t}=1]\nonumber\\
				&+\E[\E_{q_{1,m}}[\E[\psi_1|\mathbbm{1}_{1,m}=1]]],\\
				\leq&\sum_{t=2}^{T-1}\E[\psi_t|\mathbbm{1}_{t,s_t}=1](Q_t-Q_{t-1})+Q_1\E[\psi_1]\nonumber\\
				&-\E[\psi_T|\mathbbm{1}_{T,s_T}=1]-\E[\psi_1|\mathbbm{1}_{1,s_1}=1]\nonumber\\
				&+\E[\E_{q_{1,m}}[\E[\psi_1|\mathbbm{1}_{1,m}=1]]]\\
				\leq&\sum_{t=2}^{T-1}B(Q_t-Q_{t-1})+(Q_1+1)A-B-B,\\
				\leq&(Q_{T-1}-Q_1)B+(Q_1+1)A-2B,\\
				\leq&(Q_1+1)(A-B),\\
				\leq&\frac{1+\delta}{\delta}(A-B),
			\end{align}
			where we used $B\leq\psi_t\leq A$ and $1\leq Q_t\leq Q_{t-1}$.
		\end{proof}
	\end{lemma}
	
	After utilizing these results, we have the following normalized bounds, i.e., the expected regret divided by the loss range.
	\begin{theorem}\label{thm:ERST/D}
		We have the following normalized expected regret bound:
		\begin{align*}
			\frac{\E[R_{S_T}]}{D}\leq& 1+\frac{M}{\epsilon_{T}}+\sum_{t=1}^{T}\epsilon_t+{\gamma\sqrt{M\sum_t \frac{1}{\epsilon_t}}}\\
			&+\frac{(W_T+\gamma)	}{\gamma}\sqrt{M\sum_t \frac{1}{\epsilon_t}+\frac{M^2}{\epsilon_T^2}},
		\end{align*}
		where $W_T\triangleq W(\Lambda_T)$, $D\triangleq A-B$ and $\epsilon_t$ is nonincreasing.
		\begin{proof}
			The proof comes from combining the results of \autoref{thm:ERST}, \autoref{thm:EVT}, \autoref{thm:EDT} and \autoref{thm:EPsiT}.
		\end{proof}
	\end{theorem}

	Using this normalized regret bound, we can create the following cleaner bound.
	\begin{corollary}
		We have the following cleaner normalized expected regret bound:
		\begin{align}
			\frac{\E[R_{S_T}]}{D}\leq& 1+\sum_{t=1}^{T}\epsilon_t+\frac{(W_T+2\gamma)	}{\gamma}\frac{M}{\epsilon_{T}}\\
			&+\frac{(W_T+\gamma+\gamma^2)	}{\gamma}\sqrt{M\sum_t \frac{1}{\epsilon_t}},
		\end{align}
		where $W_T\triangleq W(\Lambda_T)$ and $D\triangleq A-B$.
		\begin{proof}
			The proof comes from using \autoref{thm:ERST/D} with the fact that $\sqrt{x+y}\leq \sqrt{x}+\sqrt{y}$ for all $x,y\geq 0$.
		\end{proof}
	\end{corollary}
	
	\subsection{Expected Regret Bound}
	To construct the final version of the normalized expected regret, we start by setting
	\begin{align}
		\gamma=\sqrt{W},\label{gamma}
	\end{align}
	where $W_T\leq W$ is an upper bound on the complexity.
	\begin{corollary}
		With the parameter $\gamma$ in \eqref{gamma}, we have
		\begin{align}
			\frac{\E[R_{S_T}]}{D}\leq& 1+\sum_{t=1}^{T}\epsilon_t+(\sqrt{W}+2)\frac{M}{\epsilon_{T}}\\
			&+(2\sqrt{W}+1)\sqrt{M\sum_t \frac{1}{\epsilon_t}},
		\end{align}
		where $W_T\leq W$ and $D\triangleq A-B$.
	\end{corollary}
	
	To create the final version of the expected regret bound, we design the only remaining parameter, the uniform mixture coefficients, as follows:
	\begin{align}
		\epsilon_t=\min(1,M^{\frac{1}{3}}W^{\frac{1}{3}}t^{-\frac{1}{3}}),\label{et}
	\end{align}
	which is a nonincreasing sequence of mixture weights.
	\begin{corollary}
		With the uniform mixture weights in \eqref{et}, we have
		\begin{align}
			\frac{\E[R_{S_T}]}{D}\leq&O(M^{\frac{1}{3}}W^{\frac{1}{3}}T^\frac{2}{3})
		\end{align}
		where $W_T\leq W$ and $D\triangleq A-B$.
		\begin{proof}
			From \eqref{et}, we have
			\begin{align}
				\sum_{t=1}^T\epsilon_t\leq& \sum_{t=1}^{T}M^{\frac{1}{3}}W^{\frac{1}{3}}t^{-\frac{1}{3}}\leq O(M^{\frac{1}{3}}W^{\frac{1}{3}}T^{\frac{2}{3}}),
			\end{align}
			and
			\begin{align}
				\sum_{t=1}^T\frac{1}{\epsilon_t}\leq&\frac{T}{\epsilon_{T}}\leq O(M^{-\frac{1}{3}}W^{-\frac{1}{3}}T^{\frac{4}{3}}),
			\end{align}
			which concludes the proof.
		\end{proof}
	\end{corollary}

	\begin{remark}
		Our expected regret bound
		\begin{align*}
			\E[R_{S_T}] =O\left(D{M^{\frac{1}{3}}W^{\frac{1}{3}}T^{\frac{2}{3}}}\right),
		\end{align*}
		is translation invariant and scale equivariant, hence, a fundamental regret bound \cite{cesa2007}.
	\end{remark}
	
	\section{Conclusion}
	In conclusion, we have created a completely online, generalized algorithm for the partial monitoring problem of expert advice under arbitrary limited feedback. By designing the competition class suitably, we can compete against a specific subset of the expert selection sequences in relation to the application at hand. Our normalized expected regret bounds are translation/scale-free of the expert losses and parameter free, i.e., the expected regret bound against a competition $S_T$, $\E[R_{S_T}]$, is linearly dependent on any unknown loss range $D$. Moreover, the expected regret bound is minimax optimal, i.e., $O(W^{\frac{1}{3}}T^{\frac{2}{3}})$ for a competition complexity of $W$. 
	
\bibliographystyle{ieeetran}
\bibliography{double_bib}	
	
\end{document}